\def\year{2022}\relax
\documentclass[letterpaper]{article} 
\usepackage{aaai22}  
\usepackage{times}  
\usepackage{helvet}  
\usepackage{courier}  
\usepackage[hyphens]{url}  
\usepackage{graphicx} 
\usepackage{natbib}  
\usepackage{caption} 
\DeclareCaptionStyle{ruled}{labelfont=normalfont,labelsep=colon,strut=off} 
\usepackage{amsthm,amsmath,amssymb}
\usepackage{newfloat}
\usepackage{listings}

\usepackage{wrapfig}
\usepackage{subfigure}
\usepackage{enumitem}
\usepackage{pifont}
\usepackage{float}
\usepackage{multirow}
\usepackage[ruled]{algorithm2e}
\usepackage{color}
\usepackage{hyperref}
\usepackage[normalem]{ulem}
\useunder{\uline}{\ul}{}
\usepackage{makecell}
\newcommand{\modelname}{VIB-GSL}

\usepackage[toc,title,page]{appendix}

\newtheorem{myDef}{Definition}

\newtheorem{myPro}{Proposition}
\newtheorem{myLem}{Lemma}

\usepackage{newfloat}
\usepackage{listings}
\floatstyle{ruled}
\newfloat{listing}{tb}{lst}{}
\floatname{listing}{Listing}
\title{Graph Structure Learning with Variational Information Bottleneck}
\author{
    Qingyun Sun\textsuperscript{\rm 1}\textsuperscript{\rm 2}\textsuperscript{\rm 3}, 
    Jianxin Li\textsuperscript{\rm 1}\textsuperscript{\rm 2},
    Hao Peng\textsuperscript{\rm 1},
    Jia Wu\textsuperscript{\rm 4},
    Xingcheng Fu\textsuperscript{\rm 1},
    Cheng Ji\textsuperscript{\rm 1},
    Philip S. Yu\textsuperscript{\rm 5}
}
\affiliations{
    \textsuperscript{\rm 1}
    Beijing Advanced Innovation Center for Big Data and Brain Computing,
    Beihang University,
    Beijing 100191, China\\
    \textsuperscript{\rm 2}
    School of Computer Science and Engineering, 
    Beihang University, 
    Beijing 100191, China\\
    \textsuperscript{\rm 3}
    Shenyuan Honors College,
    Beihang University,
    Beijing 100191, China\\
    \textsuperscript{\rm 4}
    School of Computing, Macquarie University, Sydney, Australia\\
    \textsuperscript{\rm 5}
    Department of Computer Science, University of Illinois at Chicago, Chicago, USA\\

    \{sunqy, lijx, penghao, fuxc, jicheng\}@act.buaa.edu.cn, jia.wu@mq.edu.au, psyu@uic.edu
}

\usepackage{bibentry}

\begin{document}

\maketitle

\begin{abstract}
Graph Neural Networks (GNNs) have shown promising results on a broad spectrum of applications. 
Most empirical studies of GNNs directly take the observed graph as input, assuming the observed structure perfectly depicts the accurate and complete relations between nodes. 
However, graphs in the real-world are inevitably noisy or incomplete, which could even exacerbate the quality of graph representations. 
In this work, we propose a novel \textbf{V}ariational \textbf{I}nformation \textbf{B}ottleneck guided \textbf{G}raph \textbf{S}tructure \textbf{L}earning framework, namely \textbf{\modelname}, in the perspective of information theory. 
\modelname~advances the Information Bottleneck (IB) principle for graph structure learning, providing a more elegant and universal framework for mining underlying task-relevant relations. 
\modelname~learns an informative and compressive graph structure to distill the actionable information for specific downstream tasks. 
\modelname~deduces a variational approximation for irregular graph data to form a tractable IB objective function, which facilitates training stability. 
Extensive experimental results demonstrate that the superior effectiveness and robustness of \modelname. 
\end{abstract}

\section{Introduction}
Recent years have seen a significant growing amount of interest in graph representation learning~\cite{zhang2018network,tong2021directed}, especially in efforts devoted to developing more effective graph neural networks (GNNs)~\cite{zhou2020graph}. 
Despite GNNs' powerful ability in learning graph representations, most of them directly take the observed graph as input, assuming the observed structure perfectly depicts the accurate and complete relations between nodes. 
However, these raw graphs are naturally admitted from network-structure data (e.g., social network) or constructed from the original feature space by some pre-defined rules, which are usually independent of the downstream tasks and lead to the gap between the raw graph and the optimal graph for specific tasks. 
Moreover, most of graphs in the real-word are noisy or incomplete due to the error-prone data collection~\cite{chen2020iterative}, which could even exacerbate the quality of representations produced by GNNs~\cite{zugner2018adversarial,sun2018adversarial}. 
It’s also found that the properties of a graph are mainly determined by some critical structures rather than the whole graph~\cite{sun2021sugar,peng2021reinforced}. 
Furthermore, many graph enhanced applications (e.g., text classification~\cite{li2020survey} and vision navigation~\cite{gao2021room}) may only have data without graph-structure and require additional graph construction to perform representation learning. 
The above issues pose a great challenge for applying GNNs to real-world applications, especially in some risk-critical scenarios. 
Therefore, \textit{learning a task-relevant graph structure is a fundamental problem for graph representation learning}. 

To adaptively learn graph structures for GNNs, many graph structure learning methods~\cite{zhu2021deep,franceschi2019learning,chen2020iterative} are proposed, most of which optimize the adjacency matrix along with the GNN parameters toward downstream tasks with assumptions (e.g., community) or certain constraints (e.g., sparsity, low-rank, and smoothness) on the graphs. 
However, these assumptions or explicit certain constraints may not be applicable to all datasets and tasks. 
There is still a lack of a general framework that can mine underlying relations from the essence of representation learning. 



Recalling the above problems, the key of structure learning problem is learning the underlying relations invariant to task-irrelevant information. 
Information Bottleneck (IB) principle~\cite{tishby2000information} provides a framework for constraining such task-irrelevant information retained at the output by trading off between prediction and compression. 
Specifically, the IB principle seeks for a representation $Z$ that is maximally informative about target $Y$ (i.e., maximize mutual information $I(Y; Z)$) while being minimally informative about input data $X$ (i.e., minimize mutual information $I(X; Z)$). 
Based on the IB principle, the learned representation is naturally more robust to data noise. 
IB has been applied to representation learning~\cite{kim2021drop,jeon2021ib,pan2020disentangled,bao2021disentangled,dubois2020learning} and numerous deep learning tasks such as model ensemble~\cite{sinha2020diversity}, fine-tuning~\cite{mahabadi2021variational}, salient region discovery~\cite{zhmoginov2019information}. 

In this paper, we advance the IB principle for graph to solve the graph structure learning problem. 
We propose a novel \textbf{V}ariational \textbf{I}nformation \textbf{B}ottleneck guided \textbf{G}raph \textbf{S}tructure \textbf{L}earning framework, namely \textbf{\modelname}. 
\modelname~employs the irrelevant feature masking and structure learning method to generate a new IB-Graph $G_{\rm IB}$ as a bottleneck to distill the actionable information for the downstream task. 
\modelname~consists of three steps: 
(1) the IB-Graph generator module learns the IB-graph ${\rm G}_{\rm IB}$ by masking irrelevant node features and learning a new graph structure based on the masked feature; 
(2) the GNN module takes the IB-graph $G_{\rm IB}$ as input and learns the distribution of graph representations; 
(3) the graph representation is sampled from the learned distribution with a reparameterization trick and then used for classification. 
The overall framework can be trained efficiently with the supervised classification loss and the distribution KL-divergence loss for the IB objective. 
The main contributions are summarized as follows: 
\begin{itemize}[leftmargin=*]
    \item 
    \modelname~advances the Information Bottleneck principle for graph structure learning, providing an elegant and universal framework in the perspective of information theory. 
    \item \modelname~is model-agnostic and has a tractable variational optimization upper bound that is easy and stable to optimize. 
    It is sufficient to plug existing GNNs into the \modelname~framework to enhance their performances. 
    \item Extensive experiment results in graph classification and graph denoising demonstrate that the proposed \modelname~enjoys superior effectiveness and robustness compared to other strong baselines. 
\end{itemize}

\section{Background and Problem Formulation}

\subsection{Graph Structure Learning}
Graph structure learning~\cite{zhu2021deep} targets jointly learning an optimized graph structure and corresponding representations to improving the robustness of GNN models. 
In this work, we focus on graph structure learning for graph-level tasks. 

Let $G\in\mathbb{G}$ be a graph with label $Y\in\mathbb{Y}$. 
Given a graph $G=(X, A)$ with node set $V$, node feature matrix $X\in\mathbb{R}^{|V|\times d}$, and adjacency matrix $A\in\mathbb{R}^{|V|\times |V|}$, or only given a feature matrix $X$, the graph structure learning problem we consider in this paper can be formulated as producing an optimized graph $G^*=(X^*, A^*)$ and its corresponding node/graph representations $Z^*=f(G^*)$, with respect to the downstream graph-level tasks. 


\subsection{Information Bottleneck}
The Information Bottleneck~\cite{tishby2000information} seeks the balance between data fit and generalization using the mutual information as both cost function and regularizer. 
We will use the following standard quantities in the information theory~\cite{cover1999elements} frequently:
Shannon entropy $H(X)=\mathbb{E}_{X\sim p(X)}[-\log p(X)]$, cross entropy $H(p(X),q(X))=\mathbb{E}_{X\sim p(X)}[-\log q(X)]$, Shannon mutual information $I(X;Y)=H(X)-H(X|Y)$, and Kullback Leiber divergence $\mathcal{D}_{\rm KL}(p(X)||q(X)=\mathbb{E}_{X\sim p(X)}\log\frac{p(X)}{q(X)}$. 
Following standard practice in the IB literature~\cite{tishby2000information}, given data $X$, representation $Z$ of $X$ and target $Y$, $(X,Y,Z)$ are following the Markov Chain $<Y\to X \to Z>$. 
\begin{myDef}[Information Bottleneck]
For the input data $X$ and its label $Y$, the \textbf{Information Bottleneck} principle aims to learn the minimal sufficient representation $Z$:
\begin{equation}
    Z = \arg\min_{Z}-I(Z; Y)+\beta I(Z; X),
\end{equation}
where $\beta$ is the Lagrangian multiplier trading off sufficiency and minimality.  
\end{myDef}

Deep VIB~\cite{alemi2016deep} proposed a variational approximation to the IB objective by parameterizing the distribution via a neural network: 
\begin{equation}
\begin{aligned}
    \mathcal{L}=\frac{1}{N}\sum^{N}_{i=1}\int &dZp(Z|X_i)\log q(Y_i|Z)\\
    &+\beta \mathcal{D}_{\rm KL}\left(p(Z|X_i),r(Z)\right),
\end{aligned}
\end{equation}
where $q(Y_i|Z)$ is the variational approximation to $p(Y_i|Z)$ and $r(Z)$ is the variational approximation of $p(Z)$. 

The IB framework has received significant attention in machine learning and deep learning~\cite{alemi2016deep,saxe2019information}. 
As for irregular graph data, there are some recent works~\cite{wu2020graph,yu2020graph, yangheterogeneous,yu2021recognizing} introducing the IB principle to graph learning. 
GIB~\cite{wu2020graph} extends the general IB to graph data with regularization of the structure and feature information for robust node representations. 
SIB~\cite{yu2020graph,yu2021recognizing} was proposed for the subgraph recognition problem. 
HGIB~\cite{yangheterogeneous} was proposed to implement the consensus hypothesis of heterogeneous information networks in an unsupervised manner. 
We illustrate the difference between related graph IB methods and our method in Section~\ref{sec:comparison}. 
\section{Variational Information Bottleneck Guided Graph Structure learning}
In this section, we elaborate the proposed \modelname, a novel variational information bottleneck principle guided graph structure learning framework. 
First, we formally define the IB-Graph and introduce a tractable upper bound for IB objective. 
Then, we introduce the graph generator to learn the optimal IB-Graph as a bottleneck and give the overall framework of \modelname. 
Lastly, we compare \modelname~with two graph IB methods to illustrate its difference and properties. 
\subsection{Graph Information Bottleneck}
\label{subsec:bound}
In this work, we focus on learning an optimal graph $G_{\rm IB}=(X_{\rm IB}, A_{\rm IB})$ named IB-Graph for $G$, which is compressed with minimum information loss in terms of $G$'s properties. 
\begin{myDef}[IB-Graph]
For a graph $G=(X,A)$ and its label $Y$, the optimal graph $G_{\rm IB}=(X_{\rm IB},A_{\rm IB})$ found by Information Bottleneck is denoted as \textbf{IB-Graph}:
\begin{equation}
\label{eq:IB-Graph}
    G_{\rm IB} = \arg\min_{G_{\rm IB}}-I(G_{\rm IB}; Y)+\beta I(G_{\rm IB}; G),
\end{equation}
where $X_{\rm IB}$ is the task-relevant feature set and $A_{\rm IB}$ is the learned task-relevant graph adjacency matrix. 
\end{myDef}


Intuitively, the first term $-I(G_{\rm IB}; Y)$ is the \textit{prediction} term, which encourages that essential information to the graph property is preserved. 
The second term $I(G_{\rm IB}; G)$ is the \textit{compression} term, which encourages that label-irrelevant information in $G$ is dropped. 
And the Lagrangian multiplier $\beta$ indicates the degree of information compression, where larger $\beta$ indicates more information in $G$ was retained to $G_{\rm IB}$. 
Suppose $G_n\in \mathbb{G}$ is a task-irrelevant nuisance in $G$, the learning procedure of $G_{\rm IB}$ follows the Markov Chain $<(Y, G_n)\to G\to G_{\rm IB}>$. 
IB-Graph only preserves the task-relevant information in the observed graph $G$ and is invariant to nuisances in data. 
\begin{myLem}[\textbf{Nuisance Invariance}]
\label{lem:1}
Given a graph $G\in\mathbb{G}$ with label $Y\in\mathbb{Y}$, let $G_n\in \mathbb{G}$ be a task-irrelevant nuisance for $Y$. 
Denote $G_{\rm IB}$ as the IB-Graph learned from $G$, then the following inequality holds: 
\begin{equation}
    I(G_{\rm IB}; G_n)\le I(G_{\rm IB}; G)-I(G_{\rm IB}; Y)
\end{equation}
\end{myLem}
Please refer to the Technical Appendix for the detailed proof. 
Lemma~\ref{lem:1} indicates that optimizing the IB objective in Eq.~\eqref{eq:IB-Graph} is equivalent to encourage $G_{\rm IB}$ to be less related to task-irrelevant information in $G$, leading to the nuisance-invariant property of IB-Graph. 

Due to the non-Euclidean nature of graph data and the intractability of mutual information, the IB objective in Eq.~\eqref{eq:IB-Graph} is hard to optimize directly. 
Therefore, we introduce two tractable variational upper bounds of $-I(G_{\rm IB}; Y)$ and $I(G_{\rm IB}; G)$, respectively. 
First, we examine the prediction term $-I(G_{\rm IB}; Y)$ in Eq.~\eqref{eq:IB-Graph}, which encourages $G_{\rm IB}$ is informative of $Y$. 
Please refer to Technical Appendix for the detailed proof of Proposition~\ref{pro:1}. 
\begin{myPro}[\textbf{Upper bound of} $-I(G_{\rm IB}; Y)$]
\label{pro:1}
For graph $G\in\mathbb{G}$ with label $Y\in\mathbb{Y}$ and IB-Graph $G_{\rm IB}$ learned from $G$, we have
\begin{equation}
\label{eq:bound1}
\begin{aligned}
    -I(Y;G_{\rm IB})\le&-\iint p(Y,G_{\rm IB})\log q_{\theta}(Y|G_{\rm IB})dYdG_{\rm IB}\\
    &+H(Y),
\end{aligned}
\end{equation}
where $q_{\theta}(Y|G_{\rm IB})$ is the variational approximation of the true posterior $p(Y|G_{\rm IB})$. 
\end{myPro}
Then we examine the compression term $I(G_{\rm IB}; G)$ in Eq.~\eqref{eq:IB-Graph}, which constrains the information that $G_{\rm IB}$ receives from $G$. 
Please refer to the Technical Appendix for the detailed proof of Proposition~\ref{pro:2}. 
\begin{myPro}[\textbf{Upper bound of} $I(G_{\rm IB}; G)$ ]
\label{pro:2}
For graph $G\in\mathcal{G}$ and IB-Graph $G_{\rm IB}$ learned from $G$, we have
\begin{equation}
\label{eq:bound2}
\begin{aligned}
    I(G_{\rm IB}; G) \le \iint p(G_{\rm IB}, G)\log \frac{p(G_{\rm IB}|G)}{r(G_{\rm IB})}dG_{\rm IB}dG,
\end{aligned}
\end{equation}
where $r(G_{\rm IB})$ is the variational approximation to the prior distribution $p(G_{\rm IB})$ of $G_{\rm IB}$. 
\end{myPro}

Finally, plug Eq.~\eqref{eq:bound1} and Eq.~\eqref{eq:bound2} into Eq.~\eqref{eq:IB-Graph} to derive the following objective function, which we try to minimize: 
\begin{equation}
\label{eq:overall_bound}
\begin{aligned}
    -&I(G_{\rm IB}; Y)+\beta I(G_{\rm IB}; G)\\
     \le &-\iint p(Y,G_{\rm IB})\log q_{\theta}(Y|G_{\rm IB})dYdG_{\rm IB} \\
    &+\beta \iint p(G_{\rm IB}, G)\log \frac{p(G_{\rm IB}|G)}{r(G_{\rm IB})}dG_{\rm IB}dG.
\end{aligned}
\end{equation}

\begin{figure*}
    \centering
    \includegraphics[width=1\linewidth]{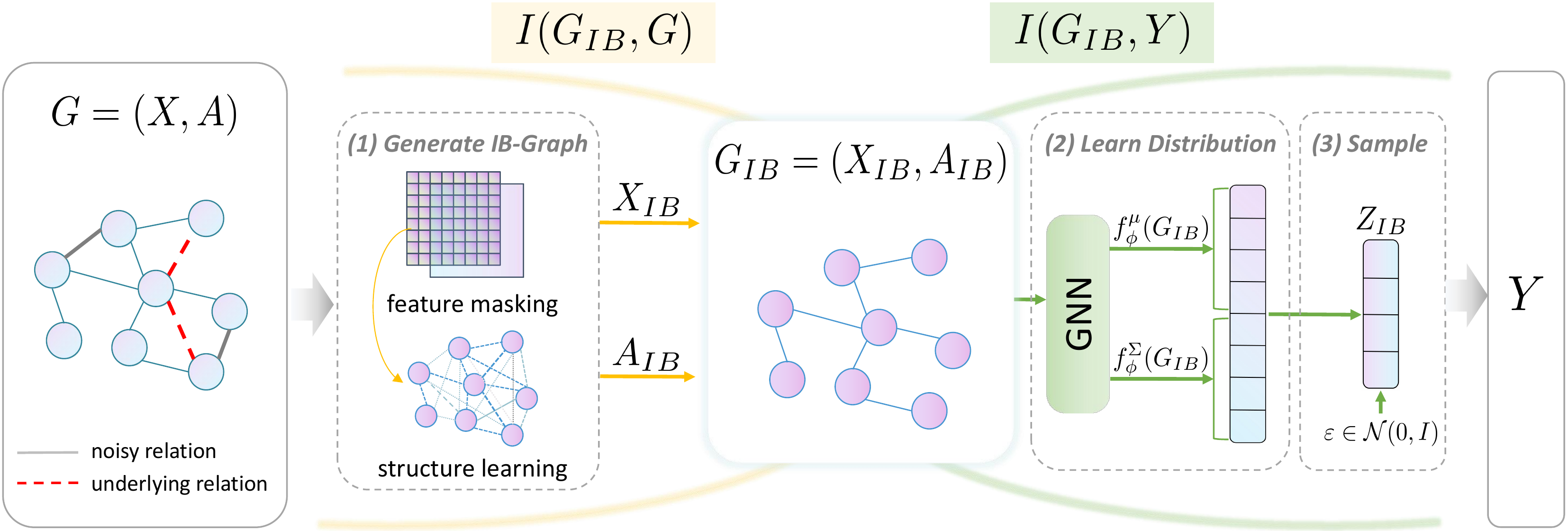}
    \caption{Overview of \modelname. 
    Given $G$ as input, \modelname~consists of the following three steps: 
    (1) Generate IB-Graph: the IB-Graph generator learns an IB-Graph $G_{\rm IB}$ by masking irrelevant features and learning a new structure; 
    (2) Learn distribution of IB-Graph representation: the GNN module learns the distribution of IB-Graph representation $Z_{\rm IB}$; 
    (3) Sample IB-Graph representation: $Z_{\rm IB}$ is sampled from the learned distribution by a reparameterization trick for classification. 
    }
    \label{fig:framework}
\end{figure*}

\subsection{Instantiating the \modelname~Framework}
\label{sec:framework}
Following the theory discussed in Section~\ref{subsec:bound}, we first obtain the graph representation $Z_{\rm IB}$ of $G_{\rm IB}$ to optimize the IB objective in Eq.~\eqref{eq:overall_bound}. 
We assume that there is no information loss during this process, which is the general practice of mutual information estimation~\cite{tian2020makes}. 
Therefore, we have $I(G_{\rm IB};Y)\approx I(Z_{\rm IB};Y)$ and $I(G_{\rm IB};G)\approx I(Z_{\rm IB};G)$.
In practice, the integral over $G_{\rm IB}$ and $G$ can be approximated by Monte Carlo sampling~\cite{shapiro2003monte} on all training samples $\{G_{i}\in \mathbb{G}, Y_{i} \in \mathbb{Y}, i=1,\dots,N\}$. 
\begin{equation}
\label{eq:overall_loss}
\begin{aligned}
    &-I(G_{\rm IB}; Y)+\beta I(G_{\rm IB}; G)
    \approx -I(Z_{\rm IB};Y) + \beta I(Z_{\rm IB};G)\\
    &\le \frac{1}{N} \! \sum^{N}_{i=1} \! \left \{ \! -\log q_{\theta}(Y_i|Z_{{\rm IB}i})
    \! + \! \beta p(Z_{{\rm IB}i}|G_i) \! \log \! \frac{p(Z_{{\rm IB}i}|G_i)}{r(Z_{\rm IB})} \! \right \}. 
\end{aligned}
\end{equation}
As shown in Figure~\ref{fig:framework}, \modelname~consists of three steps:
\subsubsection{Step-1: Generate IB-Graph $G_{\rm IB}$.}~\\
We introduce an IB-Graph generator to generate the IB-graph $G_{\rm IB}$ for the input graph $G$. 
Following the assumption that nuisance information exists in both irrelevant feature and structure, the generation procedure consists of feature masking and structure learning. 

\noindent\textbf{\textit{Feature Masking.}}
We first use a feature masking scheme to discretely drop features that are irrelevant to the downstream task, which is formulated as: 
\begin{equation}
    X_{\rm IB} = \{X_i \odot M, i=1,2,\cdots, |V|\},
\end{equation}
where $M\in\mathbb{R}^{d}$ is a learnable binary feature mask and $\odot$ is the element-wise product. 
Intuitively, if a particular feature is not relevant to task, the corresponding weight in $M$ takes value close to zero. 
We can reparameterize $X_{\rm IB}$ using the reparameterization trick~\cite{kingma2013auto} to backpropagate through a $d$-dimensional random variable:
\begin{equation}
    X_{\rm IB}=X_r+(X-X_r)\odot M,
\end{equation}
where $X_r$ is a random variable sampled from the empirical distribution of $X$. 

\noindent\textbf{\textit{Structure Learning.}}
We model all possible edges as a set of mutually independent Bernoulli random variables parameterized by the learned attention weights $\pi$: 
\begin{equation}
    A_{\rm IB}=\bigcup_{u,v \in V}\left\{a_{u,v}\sim {\rm  Ber}\left(\pi_{u,v}\right)\right\}.
\end{equation}

For each pair of nodes, we optimized the edge sampling probability $\pi$ jointly with the graph representation learning. 
$\pi_{u,v}$ describes the task-specific quality of edge $(u,v)$ and smaller $\pi_{u,v}$ indicates that  the edge $(u,v)$ is more likely to be noise and should be assigned small weight or even be removed. 
For a pair of nodes $(u,v)$, the edge sampling probability $\pi_{u,v}$ is calculated by: 
\begin{equation}
\begin{aligned}
    Z(u)&=\mathbf{NN}\left(X_{\rm IB}\left(u\right)\right),\\
    \pi_{u,v}&={\rm sigmoid}\left(Z(u)Z(v)^{\rm T}\right),
\end{aligned}
\end{equation}
where $\mathbf{NN}(\cdot)$ denotes a neural network and we use a two-layer perceptron in this work. 
One issue is that $A_{\rm IB}$ is not differentiable with respect to $\pi$ as Bernoulli distribution. 
We thus use the concrete relaxation~\cite{jang2016categorical} of the Bernoulli distribution to update $\pi$:
\begin{equation}
\label{eq:ber}
    {\rm Ber}(\pi_{u,v})\approx {\rm sigmoid}\left(\frac{1}{t}\left(\log \frac{\pi_{u,v}}{1-\pi_{u,v}} + \log \frac{\epsilon}{1-\epsilon}\right)\right),
\end{equation}
where $\epsilon \sim {\rm  Uniform}(0,1)$ and $t\in \mathbb{R}^+$ is the temperature for the concrete distribution. 
After concrete relaxation, the binary entries $a_{u,v}$ from a Bernoulli distribution are transformed into a deterministic function of $\pi_{u,v}$ and $\epsilon$. 

The graph structure after the concrete relaxation is a weighted fully connected graph, which is computationally expensive. 
We hence extract a symmetric sparse adjacency matrix by masking off those elements which are smaller than a non-negative threshold $a_0$. 
\subsubsection{Step-2: Learn Distribution of IB-Graph Representation.} 
For the compression term $I(Z_{\rm IB};G)$ in Eq.~\eqref{eq:overall_loss}, we consider a parametric Gaussian distribution as prior $r(Z_{\rm IB})$ and $p(Z_{\rm IB}|G)$ to allow an analytic computation of Kullback Leibler (KL) divergence~\cite{hershey2007approximating}:
\begin{equation}
\begin{aligned}
    r\left(Z_{\rm IB}\right)&=\mathcal{N}\left(\mu_0, \Sigma_0\right),\\
    p\left(Z_{\rm IB}|G\right)&=\mathcal{N}\left(f^{\mu}_{\phi}\left(G_{\rm IB}\right), f^{\Sigma}_{\phi}\left(G_{\rm IB}\right)\right),
\end{aligned}
\end{equation}
where $\mu\in \mathbb{R}^{K}$ and $\Sigma\in \mathbb{R}^{K\times K}$ is the mean vector and the diagonal co-variance matrix of $Z_{\rm IB}$ encoded by $f_{\phi}(G_{\rm IB})$. 
The dimensionality of $Z_{\rm IB}$ is denoted as $K$, which specifies the bottleneck size. 
We model the $f_{\phi}(G_{\rm IB})$ as a graph neural network (GNN) with weights $\phi$, where $f^{\mu}_{\phi}(G_{\rm IB})$ and $f^{\Sigma}_{\phi}(G_{\rm IB})$ are the $2K$-dimensional output value of the GNN: 
\begin{equation}
\begin{aligned}
    \forall u\in V, Z_{\rm IB}(u) &= \mathbf{GNN}\left(X_{\rm IB}, A_{\rm IB}\right),\\
    \left(f^{\mu}_{\phi}\left(G_{\rm IB}\right),  f^{\Sigma}_{\phi}\left(G_{\rm IB}\right)\right)
    &=\mathbf{Pooling}\left(\left\{Z_{\rm IB}\left(u\right), \forall u\in V\right\}\right),
\end{aligned}
\end{equation}
where the first $K$-dimension outputs encode $\mu$ and the remaining $K$-dimension outputs encode $\Sigma$ (we use a softplus transform for $f^{\Sigma}_{\phi}(G_{\rm IB})$ to ensure the non-negativity). 
We treat $r(Z_{\rm IB})$ as a fixed $d$-dimensional spherical Gaussian $r(Z_{\rm IB})=\mathcal{N}(Z_{\rm IB}|0,{\rm I})$ as in \cite{alemi2016deep}. 

\subsubsection{Step-3: Sample IB-Graph Representation.}~\\
To obtain $Z_{\rm IB}$, we can use the reparameterization trick~\cite{kingma2013auto} for gradients estimation: 
\begin{equation}
    Z_{\rm IB}=f^{\mu}_{\phi}(G_{\rm IB})+f^{\Sigma}_{\phi}(G_{\rm IB})\odot\varepsilon,
\end{equation}
where $\varepsilon\in\mathcal{N}(0,{\rm I})$ is an independent Gaussian noise and $\odot$ denotes the element-wise product. 
By using the reparameterization trick, randomness is transferred to $\varepsilon$, which does not affect the back-propagation. 
For the first term $I(Z_{\rm IB},Y)$ in Eq.~\eqref{eq:overall_loss}, $q_{\theta}(Y|Z_{\rm IB})$ outputs the label distribution of learned graph $G_{\rm IB}$ and we model it as a multi-layer perceptron classifier with parameters $\theta$. 
The multi-layer perceptron classifier takes $Z_{\rm IB}$ as input and outputs the predicted label. 

\begin{algorithm}[!t]
\caption{The overall process of \modelname}
\label{alg:algorithm}
\LinesNumbered
\KwIn{Graph $G=(X, A)$ with label $Y$; Number of training epochs $E$;}
\KwOut{IB-graph $G_{\rm IB}$, predicted label $\hat{Y}$}
Parameter initialization;\\
\For{$e=1,2,\cdots,E$}{
\tcp{Learn IB-Graph}
$X_{\rm IB} \leftarrow \{X_i \odot M, i\in |V|\}$;\\
$A_{\rm IB} \leftarrow \bigcup_{u,v \in V}\{a_{u,v}\sim {\rm  Ber}(\pi_{u,v})\}$;\\
$G_{\rm IB}  \leftarrow (X_{\rm IB}, A_{\rm IB})$;\\
\tcp{Learn distribution}
Encode $(f^{\mu}_{\phi}(G_{\rm IB}), f^{\Sigma}_{\phi}(G_{\rm IB}))$ by a GNN;\\
\tcp{Sample graph representation}
Reparameterize $Z_{\rm IB}=f^{\mu}_{\phi}(G_{\rm IB})+f^{\Sigma}_{\phi}(G_{\rm IB})\odot\varepsilon$;\\
\tcp{Optimize}
$\mathcal{L}=\mathcal{L}_{\rm CE}(Z_{\rm IB}, Y)+\beta \mathcal{D}_{\rm KL}\left (p \left (Z_{\rm IB}|G \right)||r\left (Z_{\rm IB}\right )\right )$; \\
Update model parameters to minimize $\mathcal{L}$.
}
\end{algorithm}

\subsubsection{Training Objective.}
We can efficiently compute the upper bounds in Eq.~\eqref{eq:overall_loss} on the training data samples using the gradient descent based backpropagation techniques, as illustrated in Algorithm~\ref{alg:algorithm}. 
The overall loss is:
\begin{equation}
    \mathcal{L}=\mathcal{L}_{\rm CE}(Z_{\rm IB}, Y)+\beta \mathcal{D}_{\rm KL}\left (p \left (Z_{\rm IB}|G \right)||r\left (Z_{\rm IB}\right )\right ),
\end{equation}
where $\mathcal{L}_{\rm CE}$ is the cross-entropy loss and $\mathcal{D}_{\rm KL}(\cdot||\cdot)$ is the KL divergence. 
The variational approximation proposed above facilitates the training stability effectively, as shown in Section~\ref{sec:training}. 
We also analyze the impact of compression coefficient $\beta$ on performance and learned structure in Section~\ref{sec:sensitivity}. 

\subsubsection{Property of \modelname}
Different with traditional GNNs and graph structure learning methods (e.g., IDGL~\cite{chen2020iterative}, NeuralSparse~\cite{zheng2020robust}), \modelname~is independent of the original graph structure since it learns a new graph structure. 
This property renders \modelname~extremely robust to noisy information and structure perturbations, which is verified in Section~\ref{sec:denoise}. 

\subsection{Comparison with multiple related methods.}
\label{sec:comparison}
In this subsection, we discuss the relationship between the proposed \modelname~and two related works using the IB principle for graph representation learning, i.e., GIB~\cite{wu2020graph} and SIB~\cite{yu2020graph}. 
Remark that \modelname~follows the Markov Chain $<(Y, G_n)\to G \to G_{\rm IB}>$. 


\subsubsection{\modelname~vs. GIB}
GIB~\cite{wu2020graph} aims to learn robust node representations $Z$ by the IB principle following the Markov Chain $<(Y, G_n)\to G \to Z>$. 
Specifically, GIB regularizes and controls the structure and feature information in the computation flow of latent representations layer by layer. 
Our \modelname~differs in that we aim to learn an optimal graph explicitly, which is more interpretable than denoising in the latent space. 
Besides, our \modelname~focuses on graph-level tasks while GIB focuses on node-level ones. 

\subsubsection{\modelname~vs. SIB}
SIB~\cite{yu2020graph} aims to recognise the critical subgraph $G_{sub}$ for input graph following the Markov Chain $<(Y, G_n)\to G \to G_{sub}>$. 
Our \modelname~aims to learn a new graph structure and can be applied for non-graph structured data. 
Moreover, SIB directly estimates the mutual information between subgraph and graph by MINE~\cite{belghazi2018mutual} and uses a bi-level optimization scheme for the IB objective, leading to an unstable and inefficient training process. 
Our \modelname~is more stable to train with the tractable variational approximation, which is demonstrated by experiments in Figure~\ref{fig:training}. 

\section{Experiments}

\begin{table*}[ht]
\centering
\caption{
Summary of graph classification results: “average accuracy ± standard deviation” and “improvements” (\%). \\
{\ul Underlined}: best performance of specific backbones, \textbf{bold}: best results of each dataset.
}
\label{tab:result}
\begin{tabular}{c|c|p{1.7cm}<{\centering}c|p{1.7cm}<{\centering}c|p{1.7cm}<{\centering}c|p{1.7cm}<{\centering}c}
\hline
\multirow{2}{*}{Structure Learner} & \multirow{2}{*}{Backbone} & \multicolumn{2}{c|}{IMDB-B}   & \multicolumn{2}{c|}{IMDB-M}   & \multicolumn{2}{c|}{REDDIT-B} & \multicolumn{2}{c}{COLLAB}         \\
                                   &                           & Accuracy                & $\Delta$ & Accuracy                & $\Delta$ & Accuracy         & $\Delta$        & Accuracy                & $\Delta$ \\ \hline\hline
\multirow{3}{*}{N/A}               & GCN                       & 70.7±3.7                & -        & 49.7±2.1                & -        & 73.6±4.5         & -               & 77.6±2.6                & -        \\
                                   & GAT                       & 71.3±3.5                & -        & 50.9±2.7                & -        & 73.1±2.6         & -               & 75.4±2.4                & -        \\
                                   & GIN                       & 72.1±3.8                & -        & 49.7±0.4                & -        & 85.4±3.0         & -               &  78.8±1.4               & -        \\ \hline\hline
\multirow{3}{*}{NeuralSparse}      & GCN                       & 72.0±2.6                & $\uparrow$1.3     & 50.1±3.1                & $\uparrow$0.4     & 72.1±5.2         & $\downarrow$1.5         & 76.0±2.0                & $\downarrow$1.6     \\
                                   & GAT                       & 73.4±2.2                & $\uparrow$2.1     & 53.7±3.1                & $\uparrow$2.8     & 74.3±3.1         & $\uparrow$1.2           & 75.4±5.8                & 0.0      \\
                                   & GIN                       & 73.8±1.6                & $\uparrow$1.7     & 54.2±5.4                & $\uparrow$4.5     & 86.2±2.7         & $\uparrow$0.8           & 76.6±2.1                & $\downarrow$2.2     \\ \hline
\multirow{3}{*}{SIB}       & GCN                       & 72.2±3.9                & $\uparrow$1.5     & 51.8±3.9                & $\uparrow$2.1     & 76.7±3.0         & $\uparrow$3.1           & 76.3±2.3               & $\downarrow$1.3        \\
                                   & GAT                       & 72.9±4.6                & $\uparrow$1.6     & 51.3±2.4                & $\uparrow$0.4     & 75.3±4.7         & $\uparrow$2.2           & 77.3±1.9               & $\uparrow$1.9         \\
                                   & GIN                       & 73.7±7.0                & $\uparrow$1.6     & 51.6±4.8                & $\uparrow$1.9     & 85.7±3.5         & $\uparrow$0.3           & 77.2±2.3               & $\downarrow$1.6         \\ \hline
\multirow{3}{*}{IDGL}              & GCN                       & 72.2±4.2                & $\uparrow$1.5     & 52.1±2.4                & $\uparrow$2.4     & 75.1±1.4         & $\uparrow$1.5           & 78.1±2.1               & $\uparrow$0.5     \\
                                   & GAT                       & 71.5±4.6                & $\uparrow$0.2     & 51.8±2.4                & $\uparrow$0.9     & 76.2±2.5         & $\uparrow$3.1           & 76.8±4.4               & $\uparrow$1.4     \\
                                   & GIN                       & 74.1±3.2                & $\uparrow$2.0     & 51.1±2.1                & $\uparrow$1.4     & 85.7±3.5         & $\uparrow$0.3           & 76.7±3.8               & $\downarrow$2.1     \\ \hline
\multirow{3}{*}{\modelname}        & GCN                       & {\ul 74.1±3.3}          & $\uparrow$3.4     & {\ul 54.3±1.7}          & $\uparrow$4.6     & {\ul 77.5±2.4}   & $\uparrow$3.9           & {\ul 78.3±1.4}         & $\uparrow$0.7     \\
                                   & GAT                       & {\ul 75.2±2.7}          & $\uparrow$3.9     & {\ul 54.1±2.7}          & $\uparrow$3.2     & {\ul 78.1±2.5}         & $\uparrow$5.0           & {\ul 79.1±1.2}         & $\uparrow$3.7     \\
                                   & GIN                       & {\ul \textbf{77.1±1.4}} & $\uparrow$\textbf{5.0}     & {\ul \textbf{55.6±2.0}} & $\uparrow$\textbf{5.9}    & {\ul \textbf{88.5±1.8}}  &$\uparrow$\textbf{3.1}  & {\ul \textbf{79.3±2.1}}   & $\uparrow$\textbf{0.5}     \\ \hline
\end{tabular}
\end{table*}

We evaluate \modelname\footnote{Code is available at \url{https://github.com/RingBDStack/VIB-GSL}. }~on two tasks: graph classification and graph denoising, to verify whether \modelname~can improve the effectiveness and robustness of graph representation learning. 
Then we analyze the impact of information compression quantitatively and qualitatively. 
\subsection{Experimental Setups}
\subsubsection{Datasets. }
We empirically perform experiments on \modelname~on four widely-used social datasets including IMDB-B, IMDB-M, REDDIT-B, and COLLAB~\cite{rossi2015network}. 
We choose the social datasets for evaluation because much noisy information may exist in social interactions. 
\subsubsection{Baselines. }
We compare the proposed \modelname~with a number of graph-level structure learning baselines, including NeuralSparse~\cite{zheng2020robust}, SIB~\cite{yu2020graph} and IDGL~\cite{chen2020iterative}, to demonstrate the effectiveness and robustness of \modelname. 
We do not include GIB in our baselines since it focuses on node-level representation learning. 
Similar with SIB~\cite{yu2020graph}, we plug various GNN backbones\footnote{We follow the protocol in \url{https://github.com/rusty1s/pytorch_geometric/tree/master/benchmark/kernel}. } into \modelname~including GCN~\cite{kipf2016semi}, GAT~\cite{velivckovic2017graph}, GIN~\cite{xu2019powerful} to see whether the \modelname~can boost the performance of graph classification or not. 
For a fair comparison, we use the mean pooling operation to obtain the graph representation and use a 2-layer perceptron as the graph classifier for all baselines. 

\noindent\textbf{Parameter Settings. }
We set both the information bottleneck size $K$ and the embedding dimension of baseline methods as 16. 
For \modelname, we set $t=0.1$ in Eq.~\eqref{eq:ber}, $a_0=0.1$ and perform hyperparameter search of $\beta\in\{10^{-1}, 10^{-2}, 10^{-3}, 10^{-4}, 10^{-5}, 10^{-6}\}$ for each dataset. 

\subsection{Results and Analysis}
\subsubsection{Graph Classification.} 
We first examine \modelname's capability of improving graph classification. 
We perform 10-fold cross-validation and report the average accuracy and the standard deviation across the 10 folds in Table~\ref{tab:result}, where $\Delta$ denotes the performance improvement for specific backbone and “–” indicates that there is no performance improvement for backbones without structure learner. 
The best results in each backbone group are underlined and the best results of each dataset are shown in bold.
As shown in Table~\ref{tab:result}, the proposed \modelname~consistently outperforms all baselines on all datasets by a large margin. 
Generally, the graph sparsification models (i.e., NeuralSparse and SIB) show only a small improvement in accuracy and even have a negative impact on performance (e.g., on COLLAB), which is because they are constrained by the observed structures without mining underlying relations. 
The performance superiority of \modelname~over different GNN backbones implies that \modelname~can learn better graph structure to improve the representation quality. 

\subsubsection{Graph Denoise. }
\label{sec:denoise}

\begin{figure}[!t]
    \centering
    \includegraphics[width=0.95\linewidth]{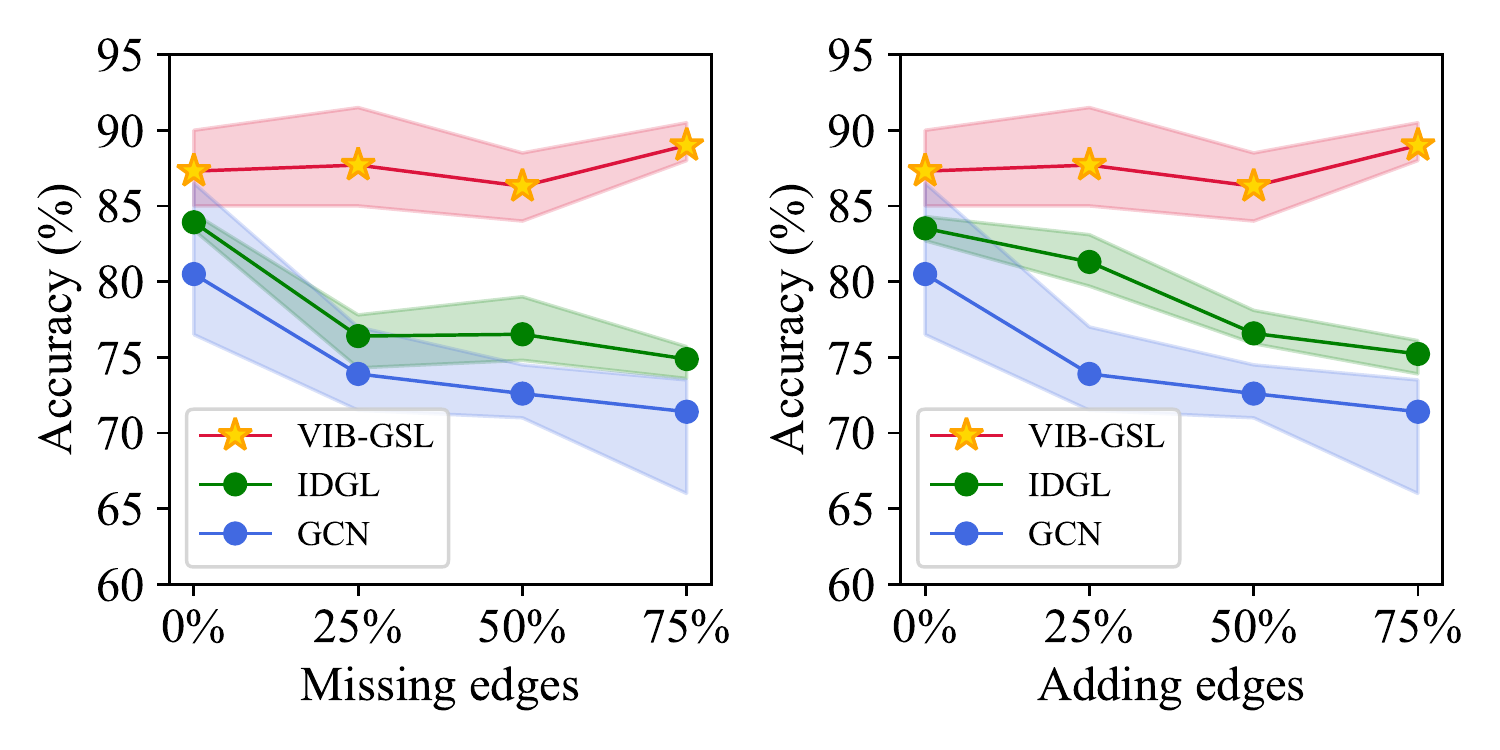}
    \caption{Test accuracy (± standard deviation) in percent for the edge attack scenarios on REDDIT-B. 
    }
    \label{fig:denoise}
\end{figure}

To evaluate the robustness of \modelname, we generate a synthetics dataset by deleting or adding edges on REDDIT-B. 
Specifically, for each graph in the dataset, we randomly remove (if edges exist) or add (if no such edges) 25\%, 50\%, 75\% edges. 
The reported results are the mean accuracy (solid lines) and standard deviation (shaded region) over 5 runs. 
As shown in Figure~\ref{fig:denoise}, the classification accuracy of GCN dropped by 5\% with 25\% missing edges and dropped by 10\% with 25\% noisy edges, indicating that GNNs are indeed sensitive to structure noise. 
Since the proposed \modelname~does not depend on the original graph structure, it achieves better results without performance degradation. 
IDGL is still sensitive to structure noise since it iteratively updates graph structure based on node embeddings, which is tightly dependent on the observed structure. 

\subsubsection{Parameter Sensitivity: Trade Off between Prediction and Compression.}
\label{sec:sensitivity}
\begin{figure}[!t]
{
\includegraphics[width=0.95\linewidth]{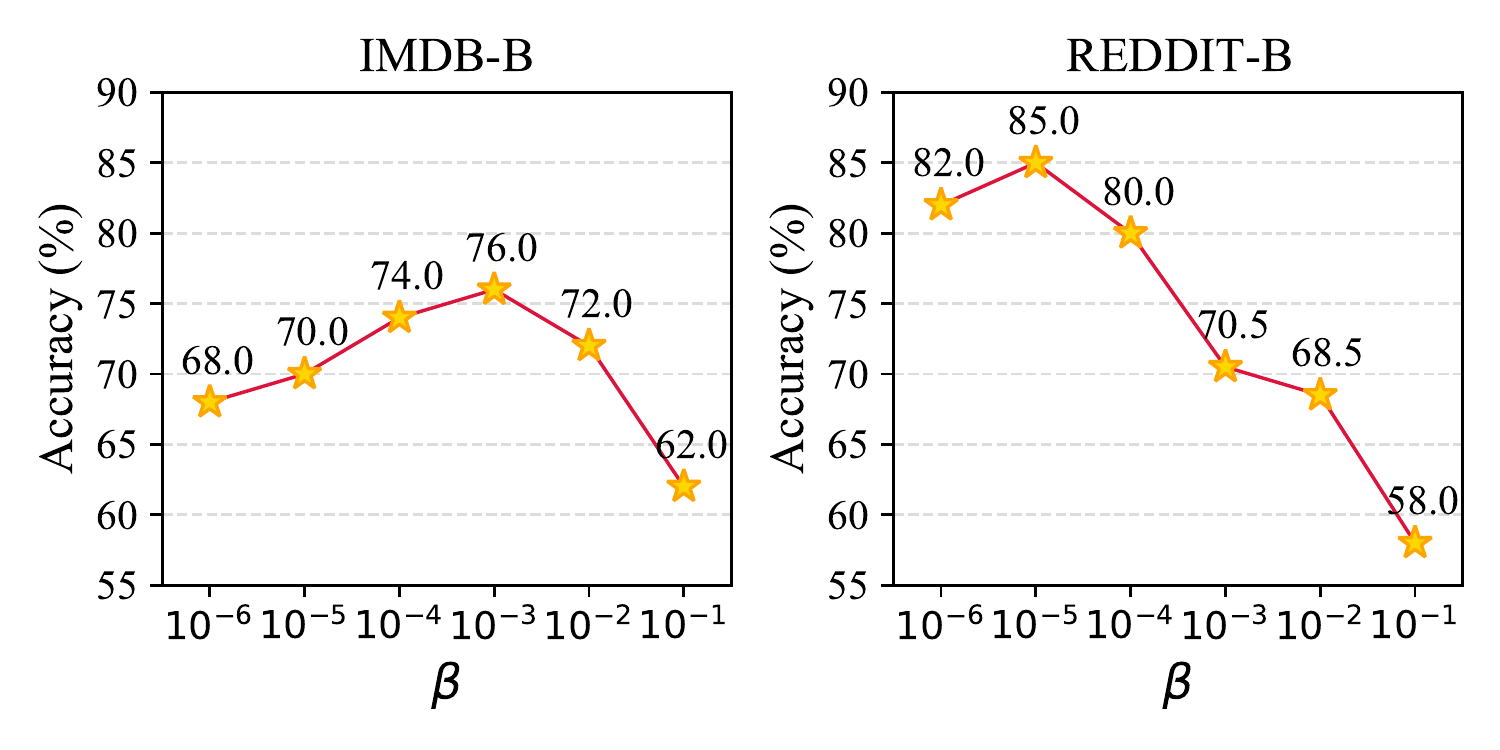}
}
\caption{Impact of $\beta$ on IMDB-B and REDDIT-B.}
\label{fig:beta}
\end{figure}
We explore the influence of the Lagrangian multiplier $\beta$ trading off prediction and compression in Eq.~\eqref{eq:IB-Graph} and Eq.~\eqref{eq:overall_loss}. 
Note that there is a relationship between increasing $\beta$ and decreasing $K$~\cite{shamir2010learning}, and the following analysis is with $K=16$. 
Figure~\ref{fig:beta} depicts the changing trend of graph classification accuracy on IMDB-B and REDDIT-B. 
Based on the results, we make the following observations: 
(1) 
Remarkably, the graph classification accuracies of \modelname~variation across different $\beta$ collapsed onto a hunchback shape on both datasets. 
The accuracy first increases with the increase of $\beta$, indicating that removing irrelevant information indeed enhances the graph representation learning. 
Then the accuracy progressively decreases and reaches very low values, indicating that excessive information compression will lose effective information. 
(2) 
Appropriate value of $\beta$ can greatly increase the model’s performance. 
\modelname~achieves the best balance of prediction and compression with $\beta=10^{-3}$ and $\beta=10^{-5}$ on IMDB-B and REDDIT-B, respectively. 
This indicates that different dataset consists of different percent of task-irrelevant information and hence needs a different degree of information compression. 


\begin{figure*}[!ht]
    \centering
    \includegraphics[width=1\linewidth]{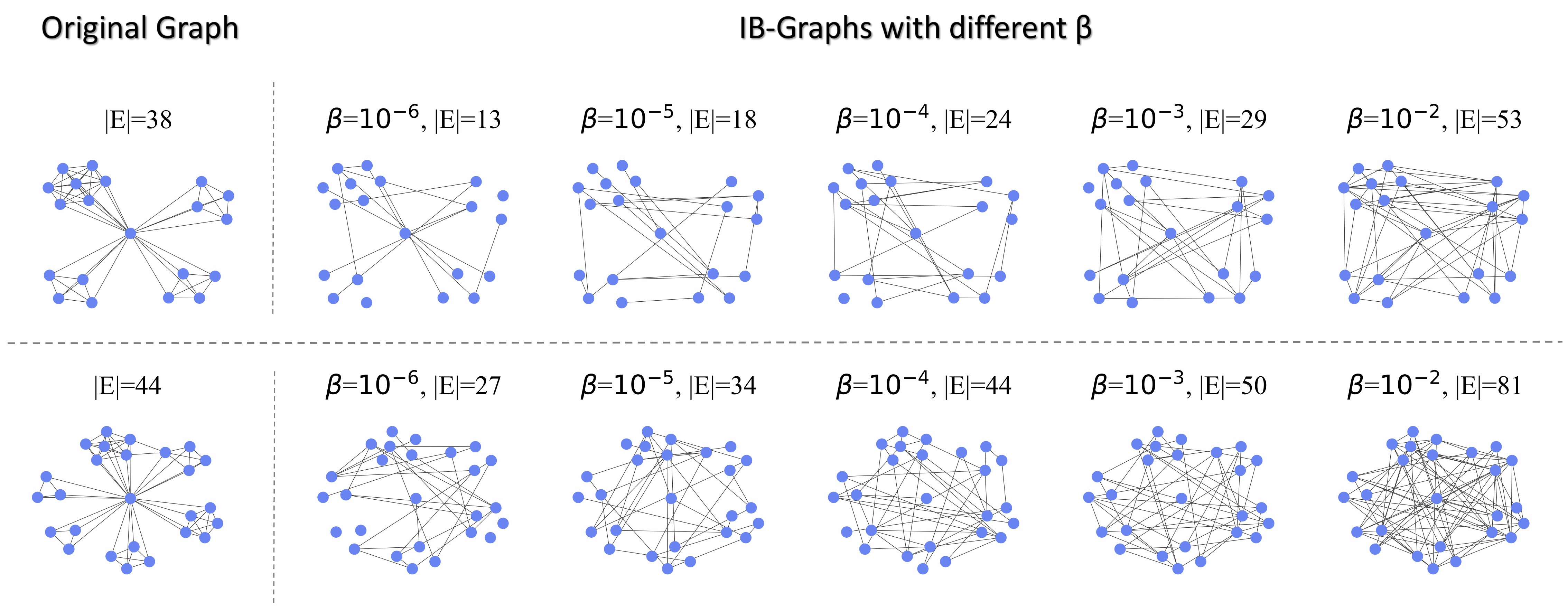}
    \caption{Original graph and IB-Graphs with different $\beta$ when \modelname~achieves the same testing performance.}
    \label{fig:visual}
\end{figure*}

\subsubsection{Graph Visualization.}
To examine the graph structure changes brought by \modelname~intuitively, we present two samples from the IMDB-B dataset and visualize the original graph and IB-Graphs learned by \modelname~in Figure~\ref{fig:visual}, where $|E|$ indicates the number of edges. 
To further analyze the impact of information compression degree, we visualize the learned IB-Graph with different $\beta$ when \modelname~achieves the same testing performance. 
Note that \modelname~does not set sparsity constraint as in most structure learning methods. 
As shown in Figure~\ref{fig:visual}, we make the following observations: 
(1)\modelname~tends to generate edges that connect nodes playing the same structure roles, which is consistent with the homophily assumption. 
(2)When achieving the same testing performance, \modelname~with larger $\beta$ will generate a more dense graph structure. 
It is because with the degree of information compression increasing, the nodes need more neighbors to obtain enough information. 

\subsubsection{Training Stability.}
\label{sec:training}


\begin{figure}[!t]
\centering
\subfigure[\modelname.]{
\label{fig:training:VIB}
\includegraphics[width=0.95\linewidth]{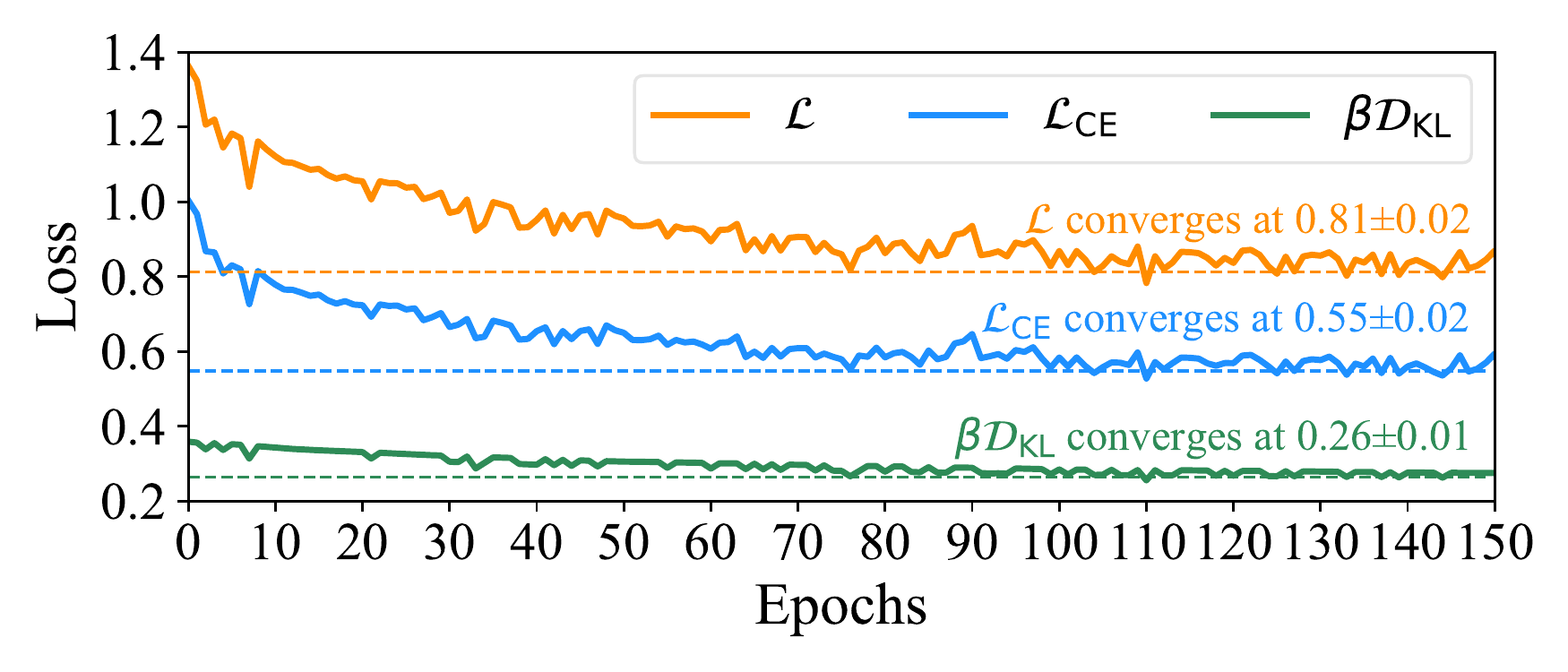}
\vspace{-0.6em}
}
\quad
\vspace{-0.6em}
\subfigure[SIB.]{
\label{fig:training:subgraph}
\includegraphics[width=0.95\linewidth]{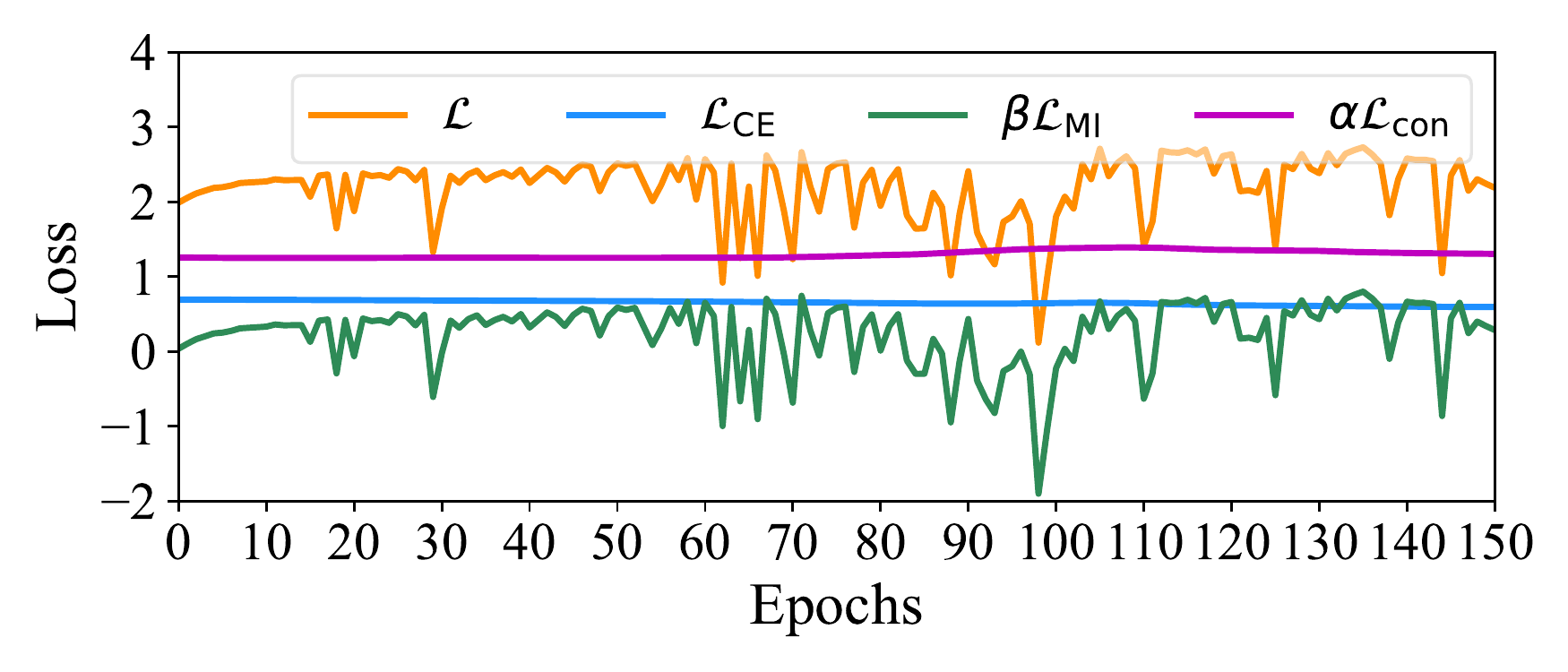}
}
\vspace{-0.6em}
\caption{Training dynamics of \modelname~and SIB.}
\label{fig:training}
\end{figure}
\vspace{-0.4em}
As mentioned in Section~\ref{sec:comparison}, \modelname~deduces a tractable variational approximation for the IB objective, which facilitates the training stability. 
In this subsection, we analyze the convergence of \modelname~and SIB~\cite{yu2020graph} on REDDIT-B with a learning rate of 0.001. 
The IB objective in~\cite{yu2020graph} is $\mathcal{L}=\mathcal{L}_{\rm CE}+\beta\mathcal{L}_{\rm MI}+\alpha\mathcal{L}_{con}$, where $\mathcal{L}_{\rm CE}$ is the cross-entropy loss, $\mathcal{L}_{\rm MI}$ is the MINE loss of estimating mutual information between original graph and learned subgraph and $\mathcal{L}_{con}$ is a connectivity regularizer. 
Figure~\ref{fig:training:VIB} depicts the losses of \modelname~(i.e., overall loss $\mathcal{L}$, cross-entropy loss $\mathcal{L}_{\rm CE}$ for classification, and the KL-divergence loss $\mathcal{D}_{\rm KL}$) with $\beta=10^{-3}$, where the dash lines indicates the mean value in the last 10 epochs when \modelname~converges. 
As mentioned in Section~\ref{sec:comparison}, SIB adopted a bi-level optimization scheme for IB objective. 
Figure~\ref{fig:training:subgraph} depicts the losses of SIB~(i.e., overall loss $\mathcal{L}$, classification loss $\mathcal{L}_{\rm CE}$, the MI estimation loss $\mathcal{L}_{\rm MI}$, and the connectivity loss $\mathcal{L}_{con}$) with $\beta=0.2$ and $\alpha=5$ as suggested in its source code. 
As shown in Figure~\ref{fig:training:VIB}, \modelname~converge steadily, showing the effectiveness of the variational approximation. 
As shown in Figure~\ref{fig:training:subgraph}, the MI estimation loss $\mathcal{L}_{\rm MI}$ is very unstable because of the bi-level optimization scheme, making SIB is very difficult to converge. 
\section{Conclusion}
In this paper, we advance the Information Bottleneck principle for graph structure learning and propose a framework named \modelname,~which jointly optimizes the graph structure and graph representations. 
\modelname~deduces a variational approximation to form a tractable IB objective function that facilitates training stability and efficiency. 
We evaluate the proposed \modelname~in graph classification and graph denoising. 
Experimental results verify the superior effectiveness and robustness of \modelname. 

\setcounter{secnumdepth}{0}
\section{Technical Appendix}

\setcounter{myLem}{0}  
\setcounter{myPro}{0}
\section{A. Proofs}
\subsection{Proof of Lemma~\ref{lem:1}}
\label{prof:lem1}
We first provide the proof of Lemma~\ref{lem:1} in Section~\ref{subsec:bound}.
\begin{proof}
We prove the Lemma~\ref{lem:1} following the same strategy of Proposition 3.1 in~\cite{achille2018emergence}. 
Suppose $G$ is defined by $Y$ and $G_n$, and $G_{\rm IB}$ depends on $G_n$ only through $G$. 
We can define the Markov Chain $<(Y, G_n)\to G \to G_{\rm IB}>$. 
According to the data processing inequality (DPI), we have:
\begin{equation}
\begin{aligned}
    I(G_{\rm IB}; G)&\ge I(G_{\rm IB}; Y, G_n)\\
    &=I(G_{\rm IB}; G_n)+I(G_{\rm IB}; Y|G_n)\\
    &=I(G_{\rm IB}; G_n)+H(Y|G_n)-H(Y|G_n; G_{\rm IB}).
\end{aligned}
\end{equation}
Since $G_n$ is be a task-irrelevant nuisance, it is independent with $Y$, we have $H(Y|G_n)=H(Y)$ and $H(Y|G_n; G_{\rm IB})\le H(Y|G_{\rm IB})$. 
Then
\begin{equation}
\begin{aligned}
    I(G_{\rm IB}; G)&\ge I(G_{\rm IB}; G_n)+H(Y|G_n)-H(Y|G_n; G_{\rm IB})\\
    &\ge I(G_{\rm IB}; G_n)+H(Y)-H(Y|G_{\rm IB})\\
    &=I(G_{\rm IB}; G_n)+I(G_{\rm IB};Y).
\end{aligned}
\end{equation}
Thus we obtain $I(G_{\rm IB}; G_n)\le I(G_{\rm IB}; G)-I(G_{\rm IB}; Y)$. 
\end{proof}

\subsection{Proof of Proposition~\ref{pro:1}}
\label{prof:pro1}
Then we provide the proof of Proposition~\ref{pro:1} in Section~\ref{subsec:bound}.

\begin{proof}
According to the definition of mutual information, 
\begin{equation}
\label{eq:pro1:eq1}
\begin{aligned}
    -I(Y,G_{\rm IB})&=-\iint p(Y,G_{\rm IB})\log \frac{p(Y,G_{\rm IB})}{p(Y)p(G_{\rm IB})}dYdG_{\rm IB}\\
    &=-\iint p(Y,G_{\rm IB})\log\frac{p(Y|G_{\rm IB})}{p(Y)}dYdG_{\rm IB},
\end{aligned}
\end{equation}
where $p(Y|G_{\rm IB})$ can be fully defined by the Markov Chain $<(Y, G_n)\to G \to G_{\rm IB}>$ as $p(Y|G_{\rm IB}) =\int p(Y|G_{\rm IB})p(G_{\rm IB}|G)dG$. 
Since $p(Y|G_{\rm IB})$ is intractable, let $q_{\theta}(Y|G_{\rm IB})$ be the variational approximation of the true posterior $p(Y|G_{\rm IB})$. 
According to the non-negativity of Kullback Leiber divergence:
\begin{equation}
\label{eq:pro1:KL}
\begin{aligned}
    &\mathcal{D}_{\rm KL}(p(Y|G_{\rm IB})||q_{\theta}(Y|G_{\rm IB})) \ge 0 \Longrightarrow \\
   &\int p(Y|G_{\rm IB})\log p(Y|G_{\rm IB}) dY\\
    & \ge \int p(Y|G_{\rm IB})\log q_{\theta}(Y|G_{\rm IB}) dY.
\end{aligned}
\end{equation} 
Plug Eq.~\eqref{eq:pro1:eq1} into Eq.~\eqref{eq:pro1:KL}, then we have
\begin{equation}
\begin{aligned}
    -I(Y,G_{\rm IB})\le&-\iint p(Y,G_{\rm IB})\log\frac{q_{\theta}(Y|G_{\rm IB})}{p(Y)}dYdG_{\rm IB}\\
    =&-\iint p(Y,G_{\rm IB})\log q_{\theta}(Y|G_{\rm IB})dYdG_{\rm IB}\\
    &+H(Y),
\end{aligned}
\end{equation}
where $H(Y)$ is the entropy of label $Y$, which can be ignored in optimization procedure. 
\end{proof}

\subsection{Proof of Proposition~\ref{pro:2}}
\label{prof:pro2}
We next provide the proof of Proposition~\ref{pro:2} in Section~\ref{subsec:bound}.

\begin{proof}
According to the definition of mutual information, 
\begin{equation}
\label{eq:pro2:eq1}
    I(G_{\rm IB}, G)=\iint p(G_{\rm IB}, G)\log \frac{p(G_{\rm IB}|G)}{p(G_{\rm IB})}dG_{\rm IB}dG.
\end{equation}
In general, computing the distribution $p(G_{\rm IB})=\int p(G_{\rm IB}|G)p(G)dG$ is very difficult, so we use $r(G_{\rm IB})$ as the variational approximation to $p(G_{\rm IB})$. 
Since the Kullback Leiber divergence $\mathcal{D}_{\rm KL}(p(Z)||r(Z))\ge 0$, 
\begin{equation}
\label{eq:pro2:KL}
\begin{aligned}
    &\mathcal{D}_{\rm KL}(p(Z)||r(Z))\ge 0\Longrightarrow \\
    &\int p(z)\log p(z) dz \ge \int p(z)\log r(z) dz.
\end{aligned}
\end{equation} 

Plug Eq.~\eqref{eq:pro2:eq1} into Eq.~\eqref{eq:pro2:KL}, then we have
\begin{equation}
    I(G_{\rm IB}, G) \le \iint p(G_{\rm IB}, G)\log \frac{p(G_{\rm IB}|G)}{r(G_{\rm IB})}dG_{\rm IB}dG.
\end{equation}
\end{proof}

\section{B. Training Efficiency}
\label{app:training}
\vspace{+1em}
For \modelname, the cost of learning an IB-Graph is $\mathcal{O}(nd+n^2d)$ for a graph with $n$ nodes in $\mathbf{R}^{d}$, while computing graph representation costs $\mathcal{O}(n^2d+ndK)$, where $d$ is the node feature dimension and $K$ is the bottleneck size. 
If we assume that $d\approx K$ and $d\ll n$, the overall time complexity is $\mathcal{O}(Kn^2)$. 

\begin{figure}[!h]
    \centering
    \includegraphics[width=1\linewidth]{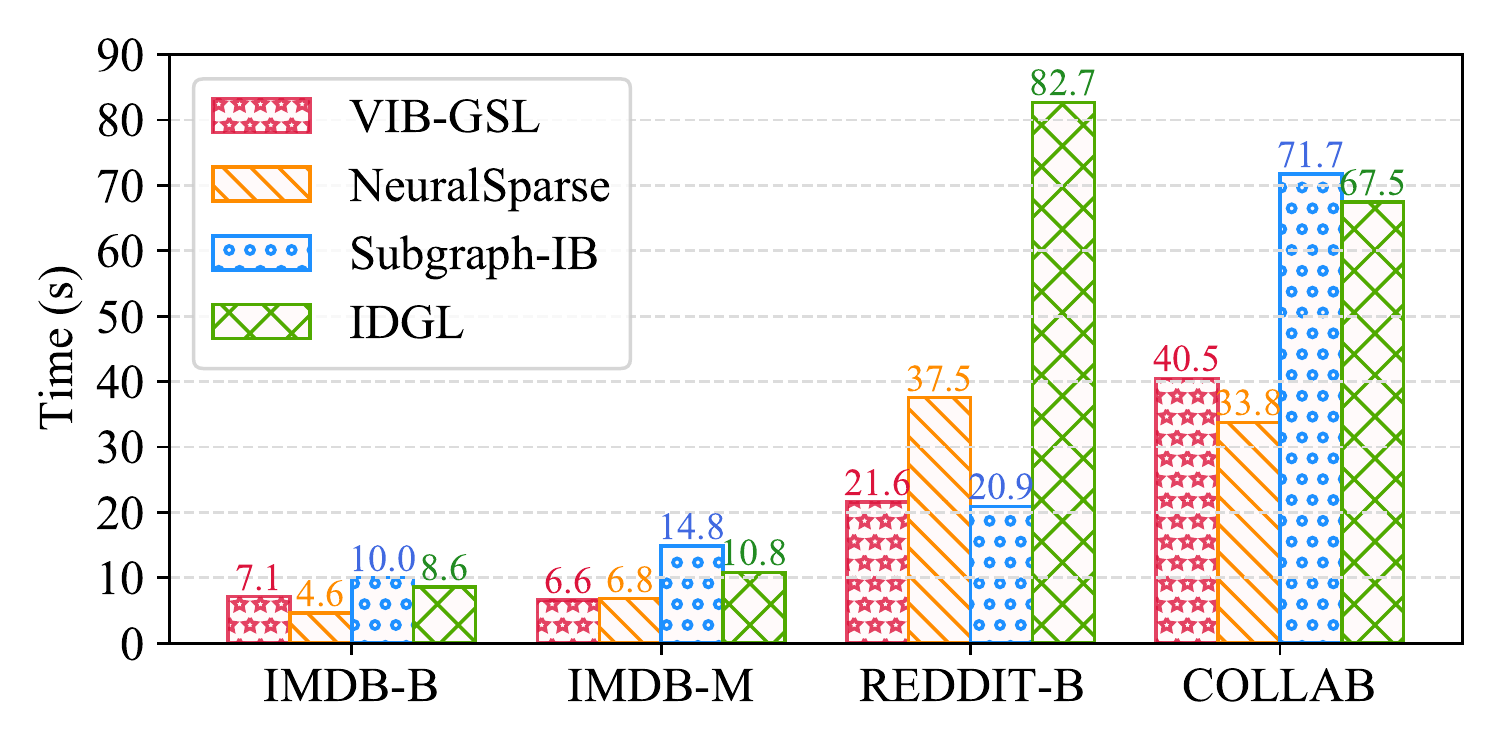}
    \caption{Training time of one epoch on various datasets.}
    \label{fig:speed}
\end{figure}

We compare the training efficiency of \modelname~with other baselines and show the mean training time of one epoch in seconds (10 runs) in Figure~\ref{fig:speed}. 
For Subgraph-IB, we set the inner loop iterations as 10. 
For IDGL, we set the maximal number of iterations in the dynamic stopping strategy to 10 as suggested in its source code. 
As shown in Figure~\ref{fig:speed}, \modelname~shows comparable efficiency with other methods when achieving the best performance. 


\clearpage
\setcounter{secnumdepth}{0}
\section{Acknowledgments}
The corresponding author is Jianxin Li. 
The authors of this paper are supported by the 
(No.U20B2053 and 61872022), 
State Key Laboratory of Software Development Environment (SKLSDE-2020ZX-12), 
Outstanding Research Project of Shen Yuan Honors College, BUAA, ( 230121208),
the ARC DECRA Project (No. DE200100964), 
and in part by NSF under grants III-1763325, III-1909323, III-2106758, and SaTC-1930941. 
 

\bibliography{ref}

\begin{thebibliography}{41}
\providecommand{\natexlab}[1]{#1}

\bibitem[{Achille and Soatto(2018)}]{achille2018emergence}
Achille, A.; and Soatto, S. 2018.
\newblock Emergence of invariance and disentanglement in deep representations.
\newblock \emph{The Journal of Machine Learning Research}, 19(1): 1947--1980.

\bibitem[{Alemi et~al.(2016)Alemi, Fischer, Dillon, and Murphy}]{alemi2016deep}
Alemi, A.~A.; Fischer, I.; Dillon, J.~V.; and Murphy, K. 2016.
\newblock Deep variational information bottleneck.
\newblock In \emph{ICLR}.

\bibitem[{Bao(2021)}]{bao2021disentangled}
Bao, F. 2021.
\newblock Disentangled Variational Information Bottleneck for Multiview
  Representation Learning.
\newblock \emph{arXiv preprint arXiv:2105.07599}.

\bibitem[{Belghazi et~al.(2018)Belghazi, Baratin, Rajeshwar, Ozair, Bengio,
  Courville, and Hjelm}]{belghazi2018mutual}
Belghazi, M.~I.; Baratin, A.; Rajeshwar, S.; Ozair, S.; Bengio, Y.; Courville,
  A.; and Hjelm, D. 2018.
\newblock Mutual information neural estimation.
\newblock In \emph{ICML}, 531--540.

\bibitem[{Chen, Wu, and Zaki(2020)}]{chen2020iterative}
Chen, Y.; Wu, L.; and Zaki, M. 2020.
\newblock Iterative deep graph learning for graph neural networks: Better and
  robust node embeddings.
\newblock In \emph{NeurIPS}.

\bibitem[{Cover(1999)}]{cover1999elements}
Cover, T.~M. 1999.
\newblock \emph{Elements of information theory}.
\newblock John Wiley \& Sons.

\bibitem[{Dubois et~al.(2020)Dubois, Kiela, Schwab, and
  Vedantam}]{dubois2020learning}
Dubois, Y.; Kiela, D.; Schwab, D.~J.; and Vedantam, R. 2020.
\newblock Learning optimal representations with the decodable information
  bottleneck.
\newblock In \emph{NeurIPS}.

\bibitem[{Franceschi et~al.(2019)Franceschi, Niepert, Pontil, and
  He}]{franceschi2019learning}
Franceschi, L.; Niepert, M.; Pontil, M.; and He, X. 2019.
\newblock Learning discrete structures for graph neural networks.
\newblock In \emph{ICML}, 1972--1982.

\bibitem[{Gao et~al.(2021)Gao, Chen, Liu, Wang, Zhang, and Wu}]{gao2021room}
Gao, C.; Chen, J.; Liu, S.; Wang, L.; Zhang, Q.; and Wu, Q. 2021.
\newblock Room-and-object aware knowledge reasoning for remote embodied
  referring expression.
\newblock In \emph{CVPR}, 3064--3073.

\bibitem[{Hershey and Olsen(2007)}]{hershey2007approximating}
Hershey, J.~R.; and Olsen, P.~A. 2007.
\newblock Approximating the Kullback Leibler divergence between Gaussian
  mixture models.
\newblock In \emph{ICASSP}, IV--317.

\bibitem[{Jang, Gu, and Poole(2017)}]{jang2016categorical}
Jang, E.; Gu, S.; and Poole, B. 2017.
\newblock Categorical reparameterization with gumbel-softmax.
\newblock In \emph{ICLR}.

\bibitem[{Jeon et~al.(2021)Jeon, Lee, Pyeon, and Kim}]{jeon2021ib}
Jeon, I.; Lee, W.; Pyeon, M.; and Kim, G. 2021.
\newblock IB-GAN: Disengangled Representation Learning with Information
  Bottleneck Generative Adversarial Networks.
\newblock In \emph{AAAI}, 7926--7934.

\bibitem[{Kim et~al.(2021)Kim, Kim, Woo, and Kim}]{kim2021drop}
Kim, J.; Kim, M.; Woo, D.; and Kim, G. 2021.
\newblock Drop-Bottleneck: Learning Discrete Compressed Representation for
  Noise-Robust Exploration.
\newblock In \emph{ICLR}.

\bibitem[{Kingma and Welling(2013)}]{kingma2013auto}
Kingma, D.~P.; and Welling, M. 2013.
\newblock Auto-encoding variational bayes.
\newblock In \emph{ICLR}.

\bibitem[{Kipf and Welling(2016)}]{kipf2016semi}
Kipf, T.~N.; and Welling, M. 2016.
\newblock Semi-Supervised Classification with Graph Convolutional Networks.
\newblock In \emph{ICLR}.

\bibitem[{Li et~al.(2020)Li, Peng, Li, Xia, Yang, Sun, Yu, and
  He}]{li2020survey}
Li, Q.; Peng, H.; Li, J.; Xia, C.; Yang, R.; Sun, L.; Yu, P.~S.; and He, L.
  2020.
\newblock A survey on text classification: From shallow to deep learning.
\newblock \emph{ACM Transactions on Intelligent Systems and Technology}.

\bibitem[{Mahabadi, Belinkov, and Henderson(2021)}]{mahabadi2021variational}
Mahabadi, R.~K.; Belinkov, Y.; and Henderson, J. 2021.
\newblock Variational Information Bottleneck for Effective Low-Resource
  Fine-Tuning.
\newblock In \emph{ICLR}.

\bibitem[{Pan et~al.(2020)Pan, Niu, Zhang, and Zhang}]{pan2020disentangled}
Pan, Z.; Niu, L.; Zhang, J.; and Zhang, L. 2020.
\newblock Disentangled Information Bottleneck.
\newblock \emph{arXiv preprint arXiv:2012.07372}.

\bibitem[{Peng et~al.(2021)Peng, Zhang, Dou, Yang, Zhang, and
  Yu}]{peng2021reinforced}
Peng, H.; Zhang, R.; Dou, Y.; Yang, R.; Zhang, J.; and Yu, P.~S. 2021.
\newblock Reinforced Neighborhood Selection Guided Multi-Relational Graph
  Neural Networks.
\newblock \emph{ACM Transactions on Information Systems}.

\bibitem[{Rossi and Ahmed(2015)}]{rossi2015network}
Rossi, R.; and Ahmed, N. 2015.
\newblock The network data repository with interactive graph analytics and
  visualization.
\newblock In \emph{AAAI}, 4292--4293.

\bibitem[{Saxe et~al.(2019)Saxe, Bansal, Dapello, Advani, Kolchinsky, Tracey,
  and Cox}]{saxe2019information}
Saxe, A.~M.; Bansal, Y.; Dapello, J.; Advani, M.; Kolchinsky, A.; Tracey,
  B.~D.; and Cox, D.~D. 2019.
\newblock On the information bottleneck theory of deep learning.
\newblock \emph{Journal of Statistical Mechanics: Theory and Experiment},
  2019(12): 124020.

\bibitem[{Shamir, Sabato, and Tishby(2010)}]{shamir2010learning}
Shamir, O.; Sabato, S.; and Tishby, N. 2010.
\newblock Learning and generalization with the information bottleneck.
\newblock \emph{Theoretical Computer Science}, 411(29-30): 2696--2711.

\bibitem[{Shapiro(2003)}]{shapiro2003monte}
Shapiro, A. 2003.
\newblock Monte Carlo sampling methods.
\newblock \emph{Handbooks in operations research and management science}, 10:
  353--425.

\bibitem[{Sinha et~al.(2020)Sinha, Bharadhwaj, Goyal, Larochelle, Garg, and
  Shkurti}]{sinha2020diversity}
Sinha, S.; Bharadhwaj, H.; Goyal, A.; Larochelle, H.; Garg, A.; and Shkurti, F.
  2020.
\newblock Diversity inducing Information Bottleneck in Model Ensembles.
\newblock In \emph{AAAI}, 9666--9674.

\bibitem[{Sun et~al.(2018)Sun, Dou, Yang, Wang, Yu, He, and
  Li}]{sun2018adversarial}
Sun, L.; Dou, Y.; Yang, C.; Wang, J.; Yu, P.~S.; He, L.; and Li, B. 2018.
\newblock Adversarial attack and defense on graph data: A survey.
\newblock \emph{arXiv preprint arXiv:1812.10528}.

\bibitem[{Sun et~al.(2021)Sun, Li, Peng, Wu, Ning, Yu, and He}]{sun2021sugar}
Sun, Q.; Li, J.; Peng, H.; Wu, J.; Ning, Y.; Yu, P.~S.; and He, L. 2021.
\newblock SUGAR: Subgraph neural network with reinforcement pooling and
  self-supervised mutual information mechanism.
\newblock In \emph{Web Conference}, 2081--2091.

\bibitem[{Tian et~al.(2020)Tian, Sun, Poole, Krishnan, Schmid, and
  Isola}]{tian2020makes}
Tian, Y.; Sun, C.; Poole, B.; Krishnan, D.; Schmid, C.; and Isola, P. 2020.
\newblock What makes for good views for contrastive learning?
\newblock In \emph{NeurIPS}.

\bibitem[{Tishby, Pereira, and Bialek(2000)}]{tishby2000information}
Tishby, N.; Pereira, F.~C.; and Bialek, W. 2000.
\newblock The information bottleneck method.
\newblock \emph{arXiv preprint physics/0004057}.

\bibitem[{Tong et~al.(2021)Tong, Liang, Ding, Dai, Li, and
  Wang}]{tong2021directed}
Tong, Z.; Liang, Y.; Ding, H.; Dai, Y.; Li, X.; and Wang, C. 2021.
\newblock Directed Graph Contrastive Learning.
\newblock \emph{Advances in Neural Information Processing Systems}, 34.

\bibitem[{Veli{\v{c}}kovi{\'c} et~al.(2017)Veli{\v{c}}kovi{\'c}, Cucurull,
  Casanova, Romero, Lio, and Bengio}]{velivckovic2017graph}
Veli{\v{c}}kovi{\'c}, P.; Cucurull, G.; Casanova, A.; Romero, A.; Lio, P.; and
  Bengio, Y. 2017.
\newblock Graph Attention Networks.
\newblock In \emph{ICLR}.

\bibitem[{Wu et~al.(2020)Wu, Ren, Li, and Leskovec}]{wu2020graph}
Wu, T.; Ren, H.; Li, P.; and Leskovec, J. 2020.
\newblock Graph information bottleneck.
\newblock In \emph{NeurIPS}.

\bibitem[{Xu et~al.(2019)Xu, Hu, Leskovec, and Jegelka}]{xu2019powerful}
Xu, K.; Hu, W.; Leskovec, J.; and Jegelka, S. 2019.
\newblock How Powerful are Graph Neural Networks?
\newblock In \emph{ICLR}.

\bibitem[{Yang et~al.(2021)Yang, Wu, Zheng, Niu, Gu, Wang, Cao, and
  Guo}]{yangheterogeneous}
Yang, L.; Wu, F.; Zheng, Z.; Niu, B.; Gu, J.; Wang, C.; Cao, X.; and Guo, Y.
  2021.
\newblock Heterogeneous Graph Information Bottleneck.
\newblock In \emph{IJCAI}, 1638--1645.

\bibitem[{Yu et~al.(2020)Yu, Xu, Rong, Bian, Huang, and He}]{yu2020graph}
Yu, J.; Xu, T.; Rong, Y.; Bian, Y.; Huang, J.; and He, R. 2020.
\newblock Graph Information Bottleneck for Subgraph Recognition.
\newblock In \emph{ICLR}.

\bibitem[{Yu et~al.(2021)Yu, Xu, Rong, Bian, Huang, and He}]{yu2021recognizing}
Yu, J.; Xu, T.; Rong, Y.; Bian, Y.; Huang, J.; and He, R. 2021.
\newblock Recognizing Predictive Substructures with Subgraph Information
  Bottleneck.
\newblock \emph{IEEE Transactions on Pattern Analysis and Machine
  Intelligence}.

\bibitem[{Zhang et~al.(2018)Zhang, Yin, Zhu, and Zhang}]{zhang2018network}
Zhang, D.; Yin, J.; Zhu, X.; and Zhang, C. 2018.
\newblock Network representation learning: A survey.
\newblock \emph{IEEE transactions on Big Data}, 6(1): 3--28.

\bibitem[{Zheng et~al.(2020)Zheng, Zong, Cheng, Song, Ni, Yu, Chen, and
  Wang}]{zheng2020robust}
Zheng, C.; Zong, B.; Cheng, W.; Song, D.; Ni, J.; Yu, W.; Chen, H.; and Wang,
  W. 2020.
\newblock Robust graph representation learning via neural sparsification.
\newblock In \emph{ICML}, 11458--11468.

\bibitem[{Zhmoginov, Fischer, and Sandler(2020)}]{zhmoginov2019information}
Zhmoginov, A.; Fischer, I.; and Sandler, M. 2020.
\newblock Information-bottleneck approach to salient region discovery.
\newblock In \emph{ECML/PKDD}, 531--546.

\bibitem[{Zhou et~al.(2020)Zhou, Cui, Hu, Zhang, Yang, Liu, Wang, Li, and
  Sun}]{zhou2020graph}
Zhou, J.; Cui, G.; Hu, S.; Zhang, Z.; Yang, C.; Liu, Z.; Wang, L.; Li, C.; and
  Sun, M. 2020.
\newblock Graph neural networks: A review of methods and applications.
\newblock \emph{AI Open}, 1: 57--81.

\bibitem[{Zhu et~al.(2021)Zhu, Xu, Zhang, Liu, Wu, and Wang}]{zhu2021deep}
Zhu, Y.; Xu, W.; Zhang, J.; Liu, Q.; Wu, S.; and Wang, L. 2021.
\newblock Deep Graph Structure Learning for Robust Representations: A Survey.
\newblock \emph{arXiv preprint arXiv:2103.03036}.

\bibitem[{Z{\"u}gner, Akbarnejad, and
  G{\"u}nnemann(2018)}]{zugner2018adversarial}
Z{\"u}gner, D.; Akbarnejad, A.; and G{\"u}nnemann, S. 2018.
\newblock Adversarial attacks on neural networks for graph data.
\newblock In \emph{ACM SIGKDD}, 2847--2856.

\end{thebibliography}
\end{document}